\def\paperID{0000}
\def\confName{CVPR}
\def\confYear{2027}

\def\paperTitle{Where Demand Meets Supply: Spatial Elastic Consensus for Low-Bandwidth Collaborative Perception}

\def\authorBlock{
    Linshan Yan \\
    School of Software, Northwestern Polytechnical University \\
    127 Youyi Xilu, Xi'an, 710072, Shaanxi, China \\
    {\tt\small yls\_nwpu@mail.nwpu.edu.cn}
}

\newif\ifreview 
\newif\ifarxiv \newcommand{\arxiv}{\arxivtrue}
\newif\ifcamera 
\newif\ifrebuttal 

\arxiv 

\pdfoutput=1
\documentclass[10pt,twocolumn,letterpaper]{article}
\ifreview \usepackage[review]{cvpr} \fi
\ifarxiv \usepackage[pagenumbers]{cvpr} \fi
\ifrebuttal \usepackage[rebuttal]{cvpr} \fi
\ifcamera \usepackage{cvpr} \fi

\usepackage{graphicx}	
\usepackage{amsmath}	
\usepackage{amssymb}	
\usepackage{booktabs}
\usepackage{times}
\usepackage{microtype}
\usepackage{epsfig}
\usepackage{caption}
\usepackage{float}
\usepackage{placeins}
\usepackage{color, colortbl}
\usepackage{stfloats}
\usepackage{enumitem}
\usepackage{tabularx}
\usepackage{xstring}
\usepackage{multirow}
\usepackage{xspace}
\usepackage{url}
\usepackage{subcaption}
\usepackage{xcolor}
\usepackage[hang,flushmargin]{footmisc}

\usepackage{algorithm}
\usepackage{algpseudocode}

\ifcamera \usepackage[accsupp]{axessibility} \fi

% ...

\ifarxiv  \fi

\newcommand{\R}[1]{{%
    \textbf{%
        \ifstrequal{#1}{1}{\textcolor{red}{R#1}}{%
        \ifstrequal{#1}{2}{\textcolor{blue}{R#1}}{%
        \ifstrequal{#1}{3}{\textcolor{magenta}{R#1}}{%
        \ifstrequal{#1}{4}{\textcolor{teal}{R#1}}{%
                           \textcolor{cyan}{R#1}%
        }}}}%
    }%
}}

\usepackage{xr-hyper}

\makeatletter
\newcommand*{\addFileDependency}[1]{
  \typeout{(#1)}
  \@addtofilelist{#1}
  \IfFileExists{#1}{}{\typeout{No file #1.}}
}

\makeatother
\newcommand*{\myexternaldocument}[1]{
    \externaldocument{#1}
    \addFileDependency{#1.tex}
    \addFileDependency{#1.aux}
}

\definecolor{cvprblue}{rgb}{0.21,0.49,0.74}
\usepackage[pagebackref,breaklinks,colorlinks,allcolors=cvprblue]{hyperref}
\usepackage[capitalize]{cleveref}
\crefname{section}{Sec.}{Secs.}
\crefname{table}{Table}{Tables}
\crefname{figure}{Fig.}{Figs.}

\ifarxiv \crefname{appendix}{App.}{Apps.}
\else \crefname{appendix}{Suppl.}{Suppls.} \fi

\unless\ifarxiv \myexternaldocument{_supplementary} \fi

\begin{document}
\title{\paperTitle}
\author{\authorBlock}
\maketitle

\begin{abstract}
Occlusions and limited sensor range make ego-only 3D perception insufficient for safe driving. Collaborative perception addresses this by enabling connected vehicles to share complementary observations, yet it remains constrained by tight vehicular communication budgets. While spatial selection is more flexible than fixed-rate feature compression, existing selectors remain largely sender-driven and waste scarce bandwidth on ego-visible targets. Furthermore, applying convolutions after masking spreads isolated tokens into their neighborhoods, silently violating low-bandwidth constraints. To address this, we propose Spatial Elastic Consensus (SEC), a bandwidth-adaptive framework centered on bilateral demand-supply consensus. Instead of unilateral token selection, the ego vehicle formulates a content-aware demand prior while each collaborator evaluates its local supply. This single-round consensus allocates the token budget to critical ego-missed evidence rather than ego-visible redundancy, without transmitting dense maps. To guarantee exact budget compliance, an anti-dilation refine-before-mask pipeline refines features prior to discrete token selection, preventing receptive-field leakage. A lightweight deformable sampler then aggregates these sparse messages with learned local offsets, compensating for the spatial context lost to token selection and mitigating inter-agent pose drift. Evaluated at a stringent 64~KiB budget, SEC consistently achieves the best AP@0.7 on the simulated OPV2V and V2XSet datasets and the real-world DAIR-V2X dataset. These results demonstrate that combining bilateral consensus with budget-exact transmission is key to efficient low-bandwidth collaborative perception.
\end{abstract}

\section{Introduction}
\label{sec:intro}
\begin{figure}[!t]
    \centering
    \includegraphics[width=\columnwidth]{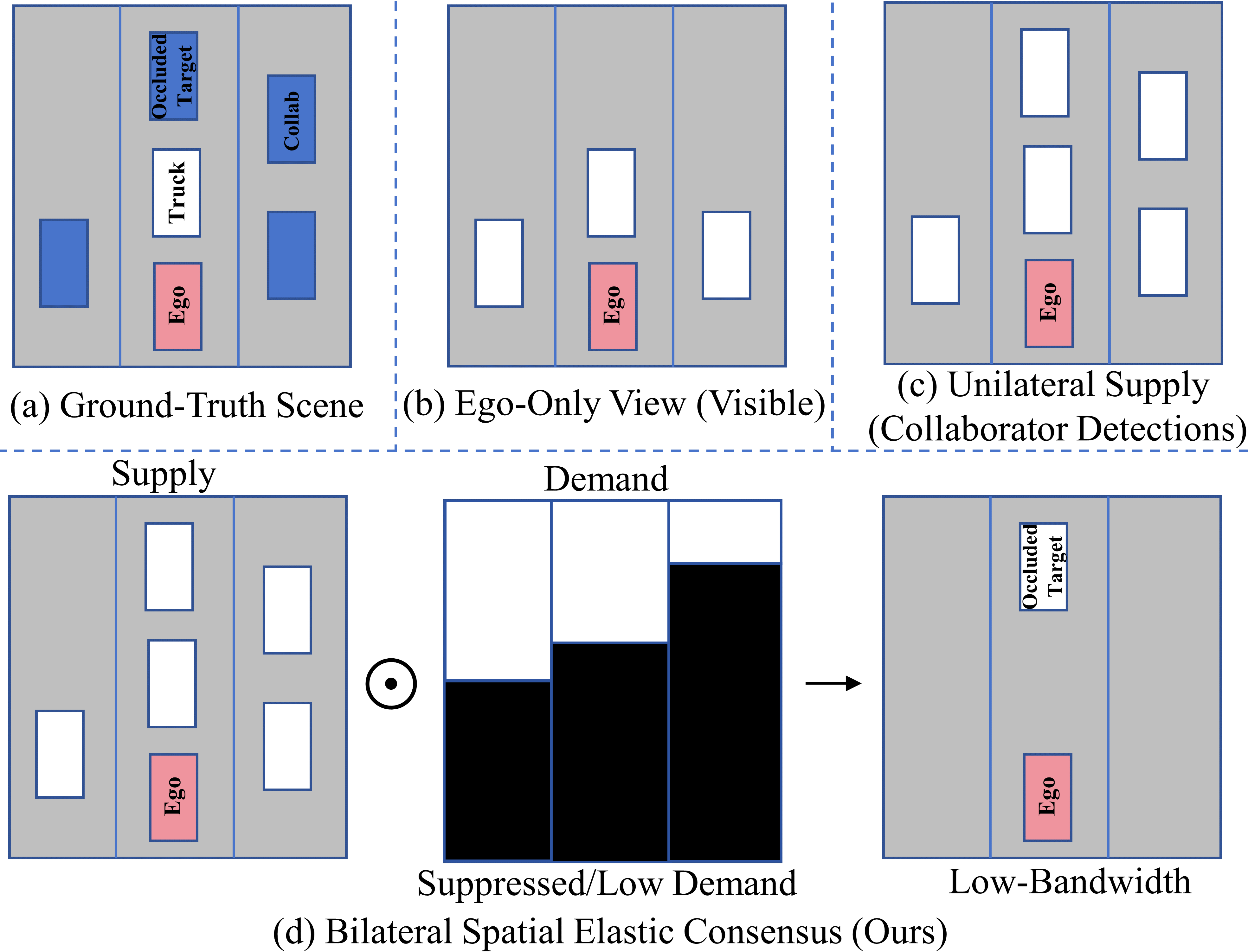}
    \caption{Concept of SEC. Top: pre-communication scene and available information. Bottom: ego demand and collaborator supply jointly select specific regions for transmission, with all other areas untransmitted.}
    \label{fig:teaser}
\end{figure}
Accurate 3D perception is a cornerstone of autonomous driving. While recent LiDAR and BEV detectors have substantially advanced single-agent 3D perception~\cite{voxelnet,pointrcnn,second,pvrcnn,centerpoint,bevdet,bevfusion}, ego-only sensing remains constrained by physical limitations such as severe occlusions and limited sensor ranges. These limitations create extensive blind spots that hide safety-critical objects, as illustrated in Fig.~\ref{fig:teaser}(a--b). Collaborative perception (CP)~\cite{proc_survey,v2xsim,fcooper} addresses this bottleneck by allowing connected agents to share intermediate bird's-eye-view (BEV) features over vehicle-to-everything (V2X) networks~\cite{opv2v,v2vnet,v2xvit,coalesce}. However, broadcasting high-dimensional BEV tensors incurs prohibitive wireless bandwidth overhead. A critical challenge for real-world CP is achieving high performance under strict per-agent communication budgets of only tens of kilobytes, which demands highly precise token selection. Unlike fixed-rate compression, token selection must decide which spatial cells carry the evidence the ego lacks; a small error in this decision wastes the entire budget. At 64~KiB per collaborator, only $0.76\%$ of the $8{,}448$ BEV cells fit in the budget.

To reduce payloads, feature compression methods encode BEV maps into compact representations~\cite{v2vnet,codefilling}, but static encoders struggle with dynamic traffic and fluctuating channel capacities. Alternatively, spatial selection methods transmit only salient spatial tokens~\cite{where2comm,select2col,who2com,when2com}. While offering greater flexibility, existing sparse selection pipelines suffer from two coupled structural limitations. First, most selectors evaluate feature confidence unilaterally~\cite{where2comm} or via decoupled scores. This sender-driven design inevitably wastes scarce bandwidth on transmitting targets the ego vehicle has already perceived, leaving critical occluded blind zones unassisted (Fig.~\ref{fig:teaser}c). Second, a pervasive architectural flaw undermines strict budget control: applying spatial convolutions after discrete token masking dilates isolated tokens into neighboring cells. This ``mask-before-refine'' process silently inflates the transmitted payload and corrupts downstream fusion, severely limiting the efficacy of ultra-low bandwidth transmission.

Reducing the payload alone does not resolve these issues: supply--demand-aware fusion~\cite{cosdh}, multi-scale deformable communication~\cite{yang2023how2comm}, and channel-wise selection~\cite{whispernet} still rely on largely sender-driven or separately computed spatial selection, and none enforces an exact $k$-token support after feature refinement. To overcome these structural limitations, effective low-bandwidth collaboration requires two coupled mechanisms: a bilateral supply--demand consensus that precisely isolates ego-missed targets, and a refine-before-mask pipeline that prevents budget inflation. Controlled diagnostics support both choices: a multiplicative demand prior shifts tokens toward ego-missed targets, and refine-before-mask keeps the support at exactly $k$ cells under identical weights.

\begin{figure*}[!t]
\centering
\includegraphics[width=\textwidth]{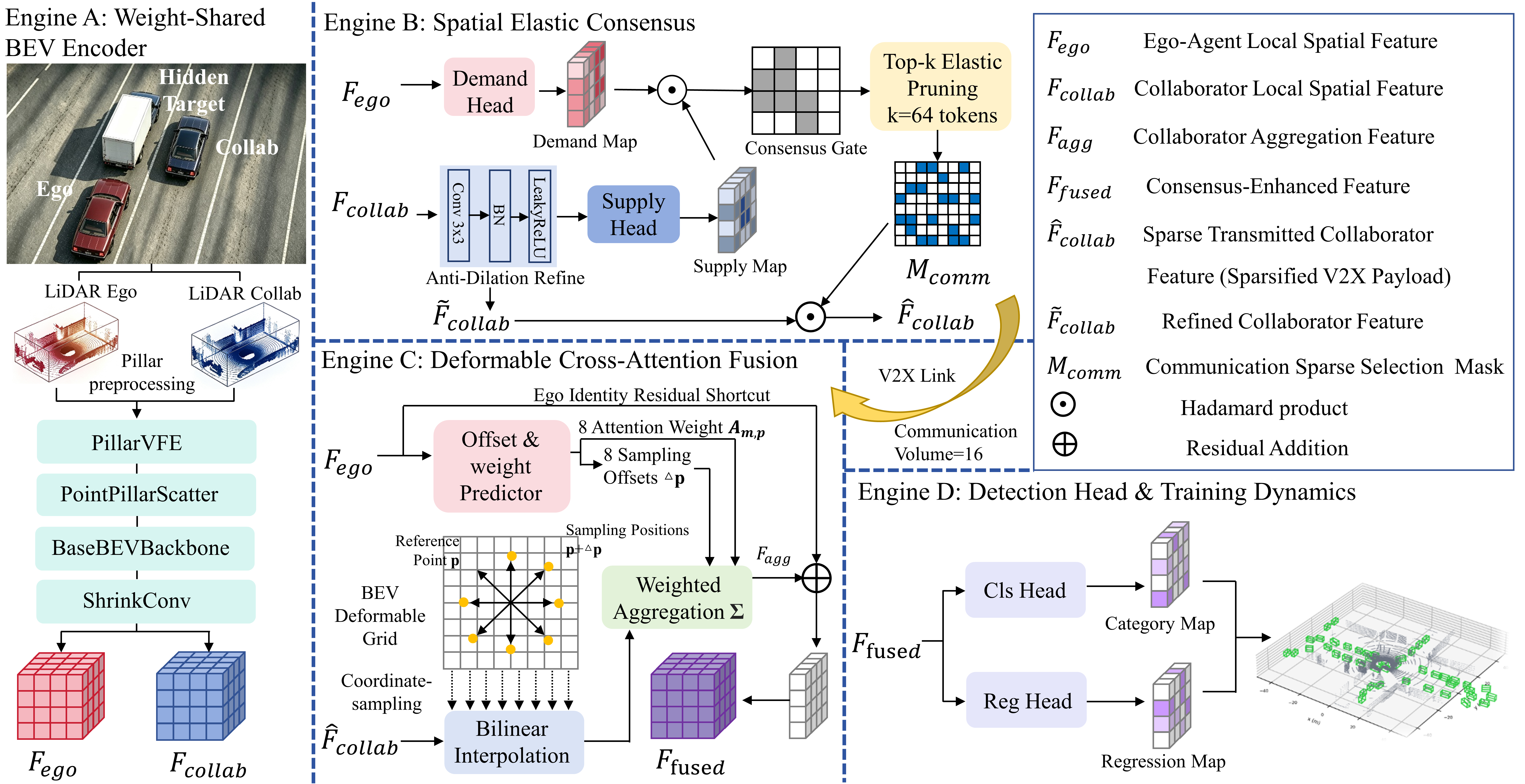}
\caption{Overall framework. Engine A extracts BEV features; Engine B selects exactly $k$ tokens ($k$: per-collaborator token budget) via bilateral consensus and refine-before-mask; Engine C aggregates them with a deformable sampler; Engine D decodes 3D boxes.}
\label{fig:framework}
\end{figure*}

We propose Spatial Elastic Consensus (SEC), a communication-efficient and budget-exact collaborative perception framework. As illustrated in Fig.~\ref{fig:framework}, SEC relies on two core designs: (1) Bilateral Consensus and Anti-Dilation Selection (Engine B). The ego agent constructs a content-aware demand map while each collaborator purifies its local features. Their single-step Hadamard product (Fig.~\ref{fig:teaser}d) isolates complementary regions and enforces exact token sparsity via a straight-through estimator with zero dilation leakage. (2) Stable Sparse Training and Aggregation (Engines C \& D). We introduce foreground warm-up and gate supervision during training to stabilize discrete selection, and employ a lightweight 8-point deformable sampler to aggregate the retained sparse tokens, effectively mitigating inter-agent pose drift.\\
Our main contributions are:
\vspace{-1mm}
\begin{itemize}
    \setlength{\itemsep}{1.5pt}
    \setlength{\parsep}{0pt}
    \item We propose SEC, a novel bandwidth-adaptive framework that introduces a single-round bilateral supply-demand consensus mechanism. It effectively suppresses ego-visible redundancy and prioritizes occluded blind spots without transmitting dense demand or supply maps.
    \item We design an anti-dilation refine-before-mask pipeline that strictly locks the transmitted payload to target token counts, eliminating post-mask receptive-field leakage and ensuring exact budget compliance at inference.
    \item Extensive experiments demonstrate that SEC achieves superior accuracy under stringent communication constraints. Notably, at a 64~KiB operating knee, SEC retains near-saturated unconstrained accuracy with only a tiny fraction of the payload.
\end{itemize}

\section{Related Work}
\label{sec:related}
\noindent\textbf{Collaborative Perception and Ego-Side Demand Priors.} Multi-agent collaborative perception (CP) overcomes single-vehicle line-of-sight limits via intermediate bird's-eye-view (BEV) feature sharing, optimally balancing representation richness and bandwidth~\cite{opv2v,v2vnet,v2xvit,dairv2x,v2xseq,v2v4real}. To address severe occlusions, prior studies quantify spatial ambiguity via Bayesian deep learning~\cite{kendall2017uncertainties}, scene completion~\cite{monoscene}, or confidence maps~\cite{where2comm}. However, these uncertainty metrics are predominantly exploited retrospectively at the receiver for feature re-weighting~\cite{coalign,hm_vit}, failing to prevent the upstream transmission of ego-visible redundancy. In contrast, we formulate a proactive ego-side saliency map (demand) to gate transmission, explicitly focusing scarce bandwidth on ego-missed regions.

\noindent\textbf{Communication-Efficient Collaboration.} To satisfy strict bandwidth constraints, early CP methods use coarse link-level handshakes~\cite{who2com,when2com}. Fine-grained methods instead prune spatial content via sender-side top-$k$ selection~\cite{where2comm}, receiver-guided queries~\cite{hm_vit,quest2024}, channel filtering~\cite{whispernet}, robust aggregation~\cite{ermvp}, codebook-based filling~\cite{codefilling}, pragmatic communication~\cite{pragmatic}, or demand-supply fusion~\cite{cosdh}. Despite this progress, two conventions persist. Selection is usually scored by one party, either the sender's own confidence or a receiver-side query, so cells the ego already observes can still consume budget and the decision never becomes joint. Refinement, in turn, typically follows masking, so the transmitted support exceeds the selected cells and no longer matches the nominal budget. SEC reverses both, forming a single agreement between demand and supply before any token is committed and moving refinement ahead of masking, so the support is exactly $k$ cells by construction. Recent work explores temporal and spatial awareness~\cite{scope2023}, cooperative reconstruction~\cite{core2023}, and contrastive view learning~\cite{collabviews}. These methods also differ in scope. Collaborator selection chooses whom to communicate with~\cite{who2com,when2com,select2col}, whereas token selection chooses what to send. They are complementary, and SEC addresses the latter.

\noindent\textbf{Sparse Deformable Alignment and Discrete Optimization.} Aggregating multi-agent features requires efficiency and resilience to pose drift. Dense Transformers capture global dependencies~\cite{v2xvit,v2xvitv2,dosovitskiy2020image,swin,bevformer,bevformerv2,bevnext} but incur severe complexity, while rigid grids fail under localization and asynchronous noise~\cite{coalign,trafalign,bevflow}. We adapt an 8-point deformable sampler from deformable attention~\cite{deformable_detr} to sparse tokens, using projections without bias terms to absorb misalignments while rejecting padding artifacts. Finally, discrete top-$k$ truncation injects gradient oscillations; a short foreground warm-up stabilizes training without changing the inference architecture.

\section{Method}
\label{sec:method}
\subsection{Overview}
\label{subsec:overview}
\noindent\textbf{Problem Formulation.}
Consider $N$ agents: one ego and $N{-}1$ collaborators. Each agent $j$ encodes its local LiDAR point cloud $\mathcal{P}^{(j)}$ into a BEV feature map via a weight-shared PointPillars backbone~\cite{pointpillars},
\begin{equation}
  F_j = \Phi_{\text{enc}}(\mathcal{P}^{(j)}) \in \mathbb{R}^{C\times H\times W},
\end{equation}
and transmits a message $\mathbf{X}_j$ under a per-frame budget of $B{=}k$ tokens. The per-collaborator budget is
\begin{equation}
  k = \min\!\left(\left\lfloor \frac{2^{V_{\text{comm}}}}{4C}\right\rfloor, HW\right),
  \label{eq:budget}
\end{equation}
where $V_{\text{comm}}$ is the bandwidth exponent and $k$ is the per-collaborator feature-token budget. The corresponding total cost for $N_c=N-1$ collaborators is
\begin{equation}
\begin{aligned}
  B_{\text{total}} &= B_{\text{feat}} + B_{\text{idx}} + B_{\text{meta}},\\
  B_{\text{feat}} &= 4 k N_c C,\qquad B_{\text{idx}}=4 k N_c,
\end{aligned}
\label{eq:overhead}
\end{equation}
where $4$ is the float32 byte width and $B_{\text{idx}}$ uses two 2-byte coordinates per token; a packed $\lceil\log_2(HW)\rceil$-bit index is smaller. For $k{=}64$, $C{=}256$, and $N_c{=}4$, the feature payload is 64~KiB per collaborator; indices add $0.39\%$, and $k{=}1024$ is the unconstrained broadcast. Hence $k$ is a direct bandwidth knob: each token costs about 1~KiB, and 64 tokens sit $16\times$ below the broadcast, with $<0.5\%$ overhead.

Let $\mathcal{C}$ be the communication policy and $\Psi$ the fusion-and-detection network. We optimize
\begin{equation}
  \max_{\mathcal{C}} \; \mathbb{E}\big[\mathrm{AP}\big(\Psi(\{\mathbf{X}_j\}_{j=0}^{N-1})\big)\big]
  \quad \text{s.t.} \quad \|\mathbf{X}_j\|_0 \le k,\;\;\forall j,
  \label{eq:problem}
\end{equation}
where $\mathrm{AP}$ is the detection AP and $j{=}0$ is the ego. The challenge is to learn $\mathcal{C}$ end-to-end while keeping the payload exactly $k$ tokens.

\noindent\textbf{Framework Overview.}
Our pipeline comprises four engines, illustrated in Fig.~\ref{fig:framework}. (1) Engine A, BEV encoding: It encodes each agent's LiDAR into a BEV feature $F_j$ with a weight-shared backbone and regroups the shared features into a multi-agent tensor together with an agent-validity mask. (2) Engine B, SEC: It identifies critical information via supply--demand interaction and packages it for low-bandwidth transmission. (3) Engine C, lightweight deformable aggregation: It aggregates the retained tokens with an ego-conditioned deformable sampler. (4) Engine D, detection and optimization: It decodes 3D boxes.

\begin{figure}[!t]
\centering
\includegraphics[width=\linewidth]{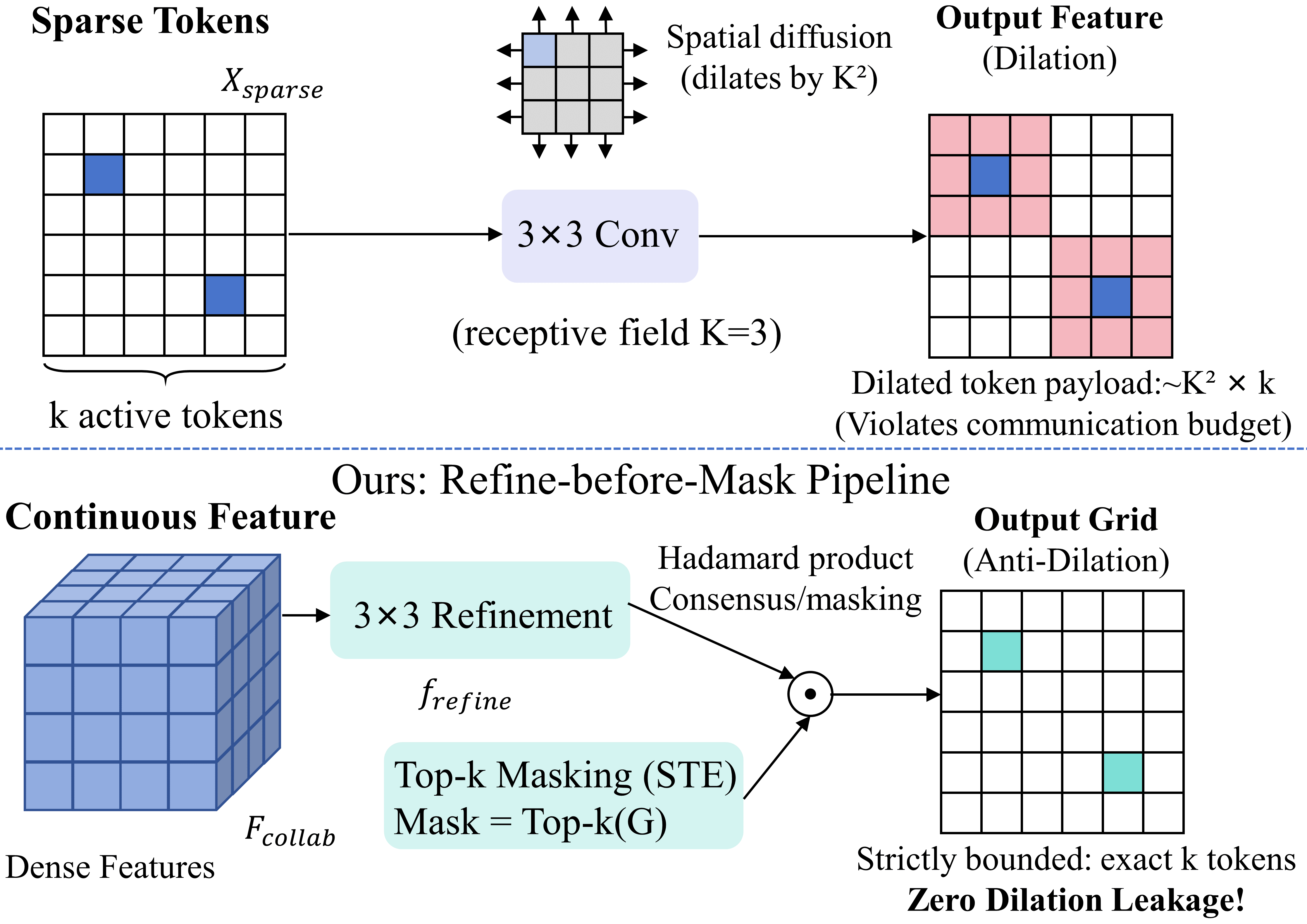}
\caption{Anti-dilation principle. Refine-before-mask keeps exactly $k$ active cells; mask-before-refine dilates each token to its neighbors and inflates the payload.}
\label{fig:antidilation-principle}
\end{figure}
\noindent\textbf{Design Rationale.}
The two bottlenecks are coupled: a sender-side selector does not know what the ego already observes, and post-mask convolution destroys the sparsity that makes token transmission affordable. SEC therefore treats selection as a joint demand-supply decision and refines features before masking, so the support stays at exactly $k$ cells: the budget is a structural property, and each kept token marks evidence the ego lacks.
\subsection{Bilateral Consensus and Anti-Dilation Selection}
\label{subsec:consensus}

\noindent\textbf{Notation.}
We write $*$ for convolution, $\odot$ for the Hadamard product, $\sigma$ for the sigmoid, $\delta$ for ReLU, BN for batch normalization, $\mathbb{I}(\cdot)$ for the indicator function, $\mathrm{Top}\text{-}k(\cdot)$ for the $k$ largest entries, $\mathrm{detach}(\cdot)$ for the stop-gradient operator, and $\|\cdot\|_0$ for the number of non-zeros; $M$ is the number of attention heads and $P$ the number of sampling points.

The role of Engine B is to decide, for every collaborator, which cells are worth transmitting under the budget. Our SEC selects what to transmit by a bilateral demand-supply agreement: the ego's demand and the collaborator's supply are combined by a Hadamard product, so a token is promoted only when both sides respond.

\noindent\textbf{Ego Demand.}
A prediction head produces an ego-side saliency map
\begin{equation}
\begin{split}
  M_{\text{demand}} = \sigma\Big(&\mathbf{W}_d^{(2)} * \delta\big(\mathrm{BN}(\mathbf{W}_d^{(1)} * F_{\text{ego}})\big)\Big)\\
  &\in (0,1)^{H\times W},
\end{split}
\label{eq:demand}
\end{equation}
The map scores each BEV location by the ego's own uncertainty and provides a content-aware prior for gating the collaborator supply. The head is shallow, so the prior is computed online without side information.

\noindent\textbf{Anti-Dilation Supply.}
A critical flaw of sparse pipelines is that a $3{\times}3$ convolution applied after masking dilates each transmitted token to its neighbours, inflating the payload by up to $9\times$. We therefore refine before selection:
\begin{equation}
  \widetilde{F}_{\text{collab}}^{(i)} = \mathrm{LeakyReLU}_{0.1}\!\big(\mathrm{BN}(\mathbf{W}_r * F_{\text{collab}}^{(i)})\big),
\end{equation}
and evaluate the collaborator supply map
\begin{equation}
\begin{split}
  M_{\text{supply}}^{(i)} = \sigma\Big(&\mathbf{W}_s^{(2)} * \delta\big(\mathrm{BN}(\mathbf{W}_s^{(1)} * \widetilde{F}_{\text{collab}}^{(i)})\big)\Big),
\end{split}
\label{eq:supply}
\end{equation}
which measures the complementary evidence available at each BEV cell. The supply head follows the same $3{\times}3$--$1{\times}1$ design as the demand head, but operates on the refined collaborator feature. Reversing the order lets the $3{\times}3$ field dilate each seed (Fig.~\ref{fig:antidilation-principle}).

\noindent\textbf{Consensus Gating.}
Transmission priority is the element-wise Hadamard product of the two maps:
\begin{equation}
  G^{(i)} = M_{\text{demand}} \odot M_{\text{supply}}^{(i)} \in (0,1)^{H\times W}.
  \label{eq:gating}
\end{equation}

The two maps need not agree: high supply with low demand means a collaborator sees what the ego already sees clearly, whereas high demand with low supply means the ego is uncertain but no collaborator holds evidence. Sender-side ranking cannot separate these two cases, because it scores cells by collaborator salience alone.

Since $G^{(i)}$ vanishes whenever either map does, a cell survives only if both sides respond; this single-round agreement is what we call consensus. The product has no learned mixing weight, and either side can remove a cell as unimportant. Because $G^{(i)}$ is computed independently per collaborator, the selector is permutation-invariant, its cost grows linearly with $N_c$, and no joint optimization across collaborators is required.

\noindent\textbf{Differentiable Elastic Selection.}
With the flattened gain $\mathbf{g}^{(i)}$, we define the hard top-$k$ mask as
\begin{equation}
\begin{split}
  \theta_k^{(i)} &= \min(\mathrm{Top}\text{-}k(\mathbf{g}^{(i)})),\\
  \mathcal{M}_{\text{hard}}^{(i)} &= \mathbb{I}(G^{(i)} \ge \theta_k^{(i)}).
\end{split}
\label{eq:hardmask}
\end{equation}
Because $\mathbb{I}(\cdot)$ is non-differentiable, we use a straight-through estimator:
\begin{equation}
  \mathcal{M}_{\text{comm}}^{(i)} = \mathcal{M}_{\text{hard}}^{(i)} + G^{(i)} - \mathrm{detach}(G^{(i)}),
  \label{eq:ste}
\end{equation}
so the forward pass transmits exactly $k$ vectors while gradients flow into both heads. The transmitted feature is $\widehat{F}_{\text{collab}}^{(i)} = \widetilde{F}_{\text{collab}}^{(i)} \odot \mathcal{M}_{\text{comm}}^{(i)}$.

\noindent\textbf{Budget-Exact Transmission.}
The forward mask $\mathcal{M}_{\text{comm}}^{(i)}$ has exactly $k$ non-zero entries, so the collaborator sends $k$ feature vectors and their coordinates; no dense map crosses the channel and refinement precedes masking, so the count cannot grow. At the receiver, selected features are regrouped with their indices and passed to Engine C, which makes the budget an architectural invariant. The bound applies to \emph{transmitted} support; the receiver may apply convolutions after transmission without increasing the number of transmitted bytes. Methods that send $k$ indices but reconstruct a dense map therefore have a different communication cost. The demand and supply heads comprise two $3{\times}3$ and two $1{\times}1$ convolutions, and the gate and mask are element-wise. At $HW{=}8448$ and $C{=}256$, the consensus engine has 0.74M parameters and takes 0.6~ms, about $7\%$ of the 9.1~ms total, while top-$k$ costs $\mathcal{O}(HW\log k)$. Selection therefore scales linearly with the number of BEV cells and introduces no pairwise interaction tensor.
\subsection{Lightweight Deformable Aggregation}
\label{subsec:fusion}

We aggregate the retained sparse tokens across agents with an 8-point deformable sampler adapted from Deformable DETR. Dense grid attention is $\mathcal{O}((HW)^2)$ and rigid grids do not absorb rotations and misalignments. Conditioned on the ego feature, we predict $P{=}8$ sampling offsets $\{\Delta\mathbf{p}_{m,s}(\mathbf{p})\}_{s=1}^{P}$ and softmax weights $\{A_{m,s}(\mathbf{p})\}_{s=1}^{P}$ per head, with $\|\Delta\mathbf{p}_{m,s}\|$ bounded by $\gamma_{\text{scale}}{=}0.1$ for local context. With value projection $\mathbf{W}_v$ (no bias, to avoid activations from padded tokens), the collaborator aggregate is
\begin{equation}
\resizebox{0.94\columnwidth}{!}{$
  F_{\text{agg}}(\mathbf{p}) = \sum_{i=1}^{N-1} v^{(i)} \sum_{m=1}^{M} \mathbf{W}_v \sum_{s=1}^{P} A_{m,s}(\mathbf{p})\, \mathcal{B}\!\left(V_m^{(i)}, \mathbf{p}+\Delta\mathbf{p}_{m,s}(\mathbf{p})\right)
$}
\label{eq:aggregation}
\end{equation}
where $V_m^{(i)}$ is the $m$-th head of $\mathbf{W}_v\widetilde{F}_{\text{collab}}^{(i)}$, $v^{(i)}$ a sample-level validity mask, and $\mathcal{B}$ bilinear grid sampling. Predicted from the ego query, the offsets target informative ego locations; the bound on $\Delta\mathbf{p}$ rules out long-range matches, and the validity mask removes empty collaborators before residual fusion. Residual fusion gives $F_{\text{fused}} = \Phi_{\text{out}}(F_{\text{ego}} + F_{\text{agg}})$, with $\Phi_{\text{out}}$ a $3{\times}3$ conv $+$ BN $+$ ReLU. Across heads the eight points specialize on different local correspondences instead of collapsing to one offset, so the sampler represents several plausible alignments at once. Reading only $P$ locations per head, its memory cost is $\mathcal{O}(HWMP)$ rather than the $\mathcal{O}((HW)^2)$ of dense cross-attention, allowing full-resolution BEV operation.

\subsection{Detection and Optimization Dynamics}
\label{subsec:dynamics}
The role of Engine D is to decode detections and stabilize the discrete selector during training. $F_{\text{fused}}$ feeds dual heads for classification and 7-DoF box regression, trained with focal/regression detection loss $\mathcal{L}_{\text{det}}$. Because the consensus heads are randomly initialized, they require supervision: we boot-strap them with an auxiliary foreground loss that supervises supply maps with ground-truth heatmaps,
\begin{equation}
\begin{split}
  \mathcal{L}_{\text{total}} &= \mathcal{L}_{\text{det}} + \lambda_{\text{aux}}(\tau)\,\mathcal{L}_{\text{aux}}, \\
  \lambda_{\text{aux}}(\tau) &= \max\!\big(0,\ \lambda_0 (1 - \tau/\tau_{\text{trans}})\big),
\end{split}
\end{equation}
where the linear decay stabilizes early training. Detection and gate-supervision losses are optimized jointly until $\tau_{\text{trans}}$; afterwards the auxiliary term is dropped and the selector is trained only through the straight-through path. This warm-up stabilizes the randomly initialized gate before the detection heads produce meaningful foreground evidence, and leaves the inference graph unchanged.

\section{Experiments}
\label{sec:experiments}
\begin{table*}[!t]
\centering
\small
\setlength{\tabcolsep}{4pt}
\caption{Performance comparison (AP@0.3 / 0.5 / 0.7) of mainstream methods.}
\label{tab:main-comparison}
\begin{tabular*}{\textwidth}{@{\extracolsep{\fill}}lcccccccccc@{}}
\toprule
& & \multicolumn{3}{c}{\textbf{OPV2V}} & \multicolumn{3}{c}{\textbf{V2XSet}} & \multicolumn{3}{c}{\textbf{DAIR-V2X}} \\
\cmidrule(lr){3-5}\cmidrule(lr){6-8}\cmidrule(lr){9-11}
\textbf{Method} & \textbf{Comm. (KiB)} & \textbf{0.3} & \textbf{0.5} & \textbf{0.7}
& \textbf{0.3} & \textbf{0.5} & \textbf{0.7} & \textbf{0.3} & \textbf{0.5} & \textbf{0.7} \\
\midrule
No Fusion   & 0      & 0.7185 & 0.7118 & 0.6136 & 0.6407 & 0.6328 & 0.5041 & 0.5669 & 0.5402 & 0.4365 \\
Late Fusion & 64     & 0.8638 & 0.8458 & 0.6885 & 0.8131 & 0.7888 & 0.6069 & 0.5267 & 0.4591 & 0.2932 \\
Early Fusion & 64     & 0.8246 & 0.8136 & 0.6818 & 0.7835 & 0.7573 & 0.5514 & 0.5806 & 0.5280 & 0.3843 \\
\midrule
DiscoNet~\cite{discoNet}     & 64     & 0.8279 & 0.7818 & 0.5669 & 0.7942 & 0.7211 & 0.4850 & 0.6087 & 0.5084 & 0.3062 \\
V2X-ViT~\cite{v2xvit}         & 64     & 0.7669 & 0.7457 & 0.5540 & 0.7078 & 0.6806 & 0.4706 & 0.5515 & 0.4966 & 0.3225 \\
CoAlign~\cite{coalign}         & 64     & 0.7408 & 0.7133 & 0.5507 & 0.6963 & 0.6458 & 0.4390 & 0.5353 & 0.4730 & 0.3591 \\
How2comm~\cite{yang2023how2comm} & 64 & 0.8583 & 0.8422 & 0.7312 & 0.8999 & 0.8693 & 0.6754 & 0.5394 & 0.4711 & 0.3150 \\
ERMVP~\cite{ermvp}           & 64     & 0.8768 & 0.8630 & 0.7536 & 0.8387 & 0.8133 & 0.6685 & 0.6016 & 0.5529 & 0.3856 \\
MRCNet~\cite{mrcnet}           & 64     & 0.8395 & 0.8220 & 0.6756 & 0.7722 & 0.7464 & 0.5388 & 0.5380 & 0.4905 & 0.3652 \\
CoSDH~\cite{cosdh}           & 64     & 0.8420 & 0.8265 & 0.6726 & 0.7684 & 0.7447 & 0.6138 & \textbf{0.6316} & 0.5700 & 0.3737 \\
MDD~\cite{huang2025v2xr}       & 64     & 0.7145 & 0.6976 & 0.5781 & 0.6490 & 0.6110 & 0.4443 & 0.5074 & 0.4540 & 0.3395 \\
WhisperNet~\cite{whispernet} & 64 & 0.8723 & 0.8671 & 0.7844 & 0.8133 & 0.8075 & 0.7079 & 0.5889 & 0.5558 & 0.4373 \\
\midrule
\textbf{Ours} & 64 & \textbf{0.9089} & \textbf{0.9034} & \textbf{0.8081} & \textbf{0.8680} & \textbf{0.8614} & \textbf{0.7461} & 0.6303 & \textbf{0.5873} & \textbf{0.4463} \\
\bottomrule
\end{tabular*}
\end{table*}

\begin{table}[t]
\centering
\small
\caption{Component-wise ablation on V2XSet at 64~KiB. $\Delta$ is relative to the full model; MBR: mask-before-refine.}
\label{tab:components}
\begin{tabular}{lcc}
\toprule
Configuration & AP@0.7 & $\Delta$ \\
\midrule
Full model (ours)                       & \textbf{0.7461} & -- \\
w/o anti-dilation (MBR, retrained)     & 0.7105 & $-0.0356$ \\
w/o 8-pt deformable (rigid grid)        & 0.6423 & $-0.1038$ \\
w/o gate supervision (none)             & 0.6594 & $-0.0867$ \\
w/ constant $\lambda_{\mathrm{aux}}$ (no decay)        & 0.7385 & $-0.0076$ \\
w/o bilateral gate (supply-only)        & 0.7435 & $-0.0026$ \\
w/o collaboration (ego-only)            & 0.5041 & $-0.2420$ \\
\midrule
Random selection (reference)            & 0.5087 & $-0.2374$ \\
\bottomrule
\end{tabular}
\end{table}

\subsection{Experimental Settings}
\label{sec:setup}
\noindent\textbf{Datasets and Evaluation Metrics.}
We evaluate on OPV2V~\cite{opv2v}, V2XSet~\cite{v2xvit}, and DAIR-V2X~\cite{dairv2x} for LiDAR-based 3D object detection, covering simulated and real-world scenarios. We use the official test splits for OPV2V and V2XSet and the validation split for DAIR-V2X, for which test annotations are not publicly released. Detection performance is measured by per-frame Average Precision (AP) at IoU thresholds 0.3, 0.5, and 0.7, without global re-sorting across frames. Communication cost is counted from the transmitted feature values and token coordinates. The main comparison fixes the feature payload at 64~KiB per collaborator. OPV2V and V2XSet scenes contain one ego vehicle and up to four collaborators within a 70~m communication range; DAIR-V2X contains one vehicle and one roadside unit.

\noindent\textbf{Implementation Details.}
We implement SEC on the OpenCOOD framework with a PointPillars backbone, a $0.4$~m BEV voxel size, at most 32 points per voxel, and $C{=}256$ feature channels. The BEV feature grid is $48\times176$ for OPV2V and V2XSet. SEC is trained end-to-end with Adam (learning rate $5\times10^{-4}$, weight decay $10^{-4}$), four warm-up epochs, and a cosine schedule. Random flip, yaw rotation, and scaling are applied consistently to all agents. All experiments run on NVIDIA RTX 4080 GPUs. At 64~KiB, SEC retains up to $99.4\%$ of the best observed AP using only $6.25\%$ of the maximum payload (Fig.~\ref{fig:supp-combined}a); we use $V_{\text{comm}}{=}16$ as the Pareto-optimal knee for all main results.

\subsection{Quantitative Evaluation}
\label{sec:main_results}
\noindent\textbf{Benchmark Comparison.}
Table~\ref{tab:main-comparison} compares SEC with prior methods at 64~KiB per collaborator; SEC attains the highest AP@0.7 on OPV2V, V2XSet, and DAIR-V2X ($0.8081$/$0.7461$/$0.4463$), beating the strongest prior method by $2.37$/$3.82$ points on OPV2V/V2XSet and $0.90$ on DAIR-V2X, and leading at AP@0.3/0.5 on OPV2V. On V2XSet, How2comm is competitive at loose thresholds ($0.8999$/$0.8693$) but falls to $0.6754$ at AP@0.7 (SEC $0.7461$), so the gap widens as IoU tightens. Over the ego-only model, SEC adds $19.5$/$24.2$ AP@0.7 points on OPV2V/V2XSet and still leads on DAIR-V2X, where margins are smaller. Table~\ref{tab:main-comparison} also shows deformable token fusion attaining the highest AP among late, early, and intermediate fusion at every threshold and dataset.

\begin{figure*}[!t]
\centering
\includegraphics[width=\textwidth]{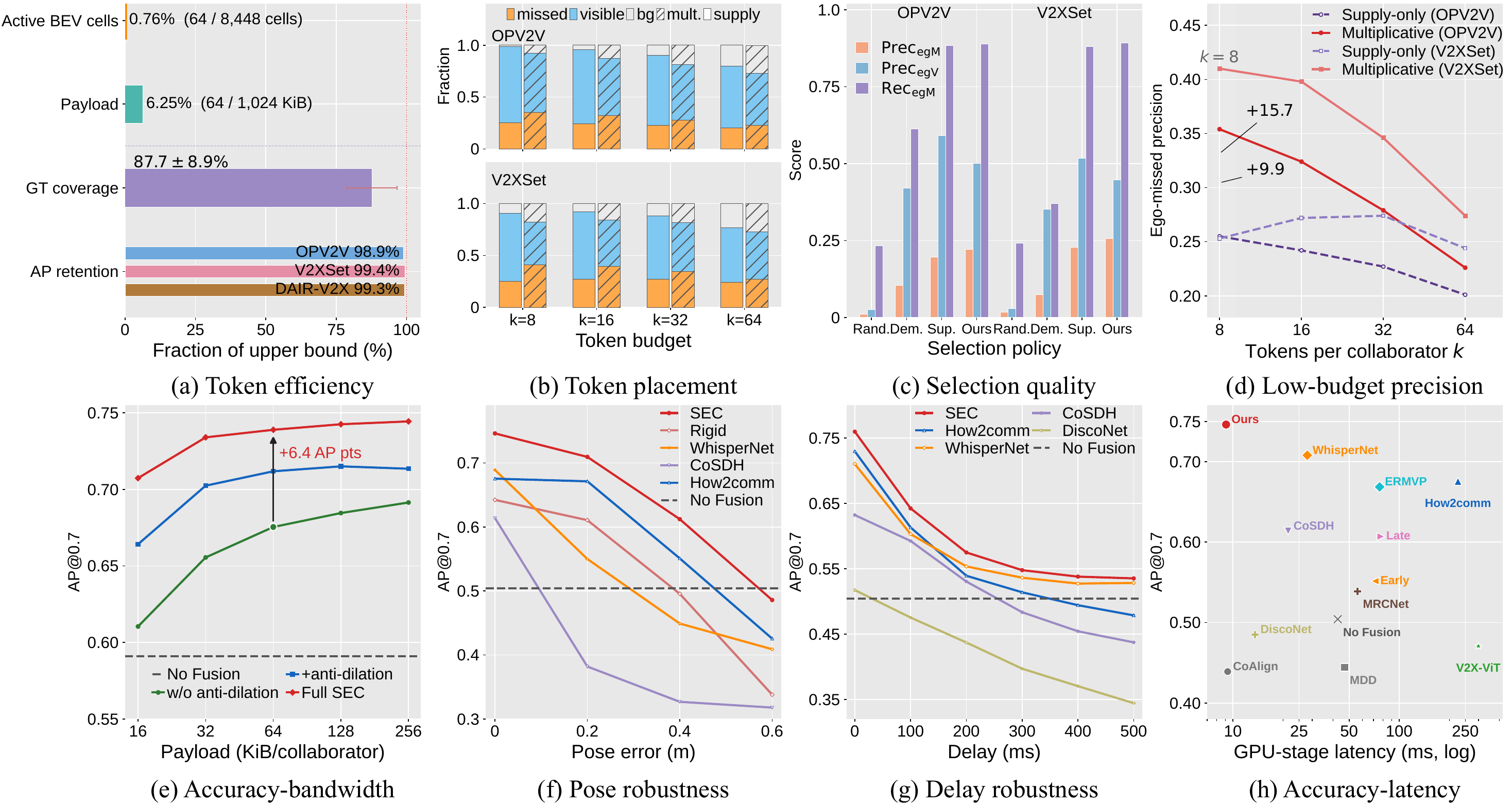}
\caption{Diagnostic analyses at 64~KiB: (a) efficiency; (b) token placement (solid: supply-only, hatched: multiplicative); (c) selection quality; (d) low-budget precision; (e)--(g) performance under bandwidth, pose, and delay; (h) accuracy--latency trade-off.}
\label{fig:supp-combined}
\end{figure*}

\noindent\textbf{Accuracy--Bandwidth Trade-off.}
Fig.~\ref{fig:supp-combined}(e) sweeps 16--256~KiB per collaborator. SEC gains $3.2$ AP@0.7 from 16 to 64~KiB but only $0.6$ for a further fourfold budget, so 64~KiB lies on the flat part of the curve. Both ablated variants rise more slowly; SEC's margin over the weaker one peaks at 16~KiB ($9.7$ points) and narrows with bandwidth.

\noindent\textbf{Robustness to Pose Noise and Communication Latency.}
Fig.~\ref{fig:supp-combined}(f) isolates the deformable sampler under pose drift: the rigid-grid variant shares the backbone and budget but uses zero offsets, and its AP@0.7 is lower at every noise level, with the gap widening at $0.6$~m/$3^\circ$. No Fusion stays flat, while WhisperNet and CoSDH degrade faster. Fig.~\ref{fig:supp-combined}(g) delays one collaborator by up to five 100~ms frames. SEC has the highest no-delay AP@0.5 ($0.898$) and stays above How2comm, WhisperNet, and CoSDH throughout; CoSDH degrades fastest and falls below no fusion at 500~ms.

\begin{figure*}[!t]
\centering
\includegraphics[width=\textwidth]{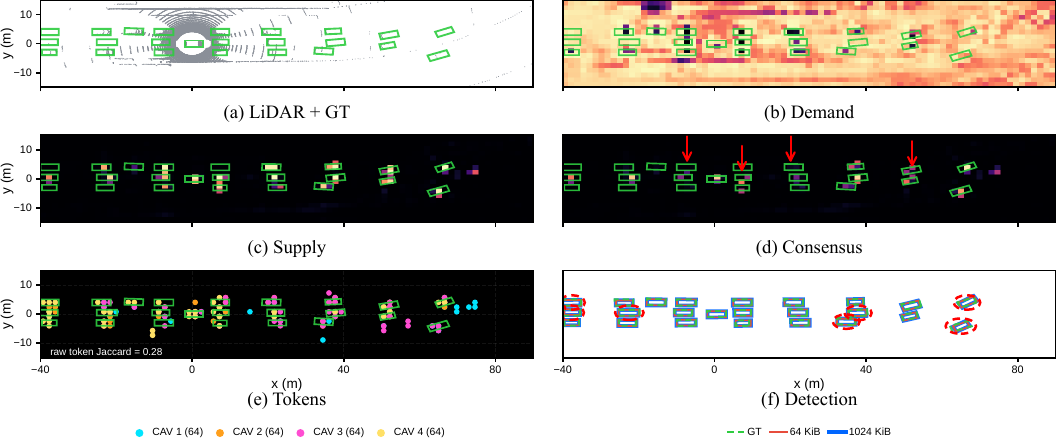}
\caption{Qualitative walk-through at $k{=}64$: (a) LiDAR/GT; (b) ego demand (where the ego needs help); (c) collaborator supply (where collaborators have evidence); (d) Hadamard (element-wise) gain; (e) selected tokens; (f) detection.}
\label{fig:qual-token}
\end{figure*}

\begin{figure}[!t]
\centering
\includegraphics[width=\linewidth]{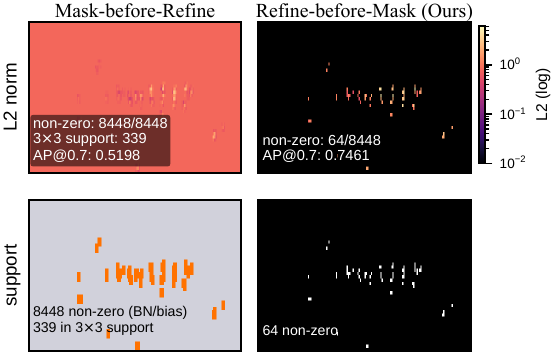}
\caption{Anti-dilation mechanism on a V2XSet frame with identical weights. Top: per-cell feature L2 norm (feature magnitude); bottom: non-zero support (active cells). Refine-before-mask keeps exactly $k$ cells; mask-before-refine becomes dense.}
\label{fig:antidilation}
\end{figure}

\begin{figure*}[!t]
\centering
\includegraphics[width=\textwidth]{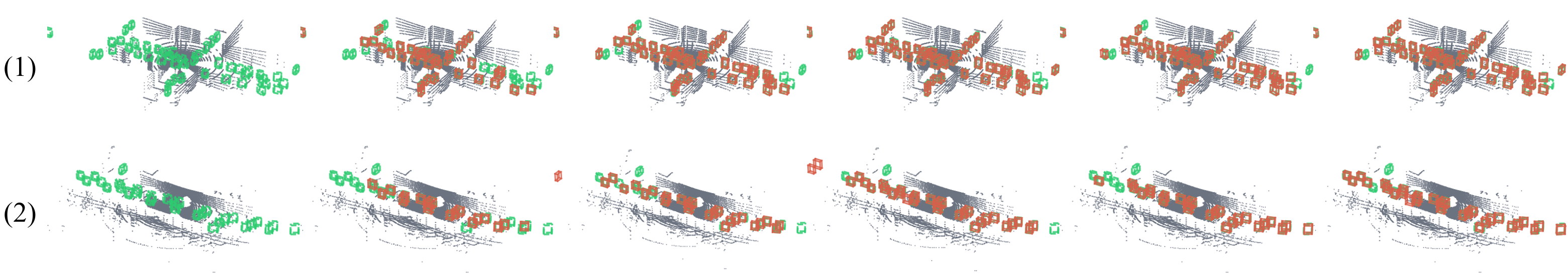}
\par\vspace{0.5pt}\noindent\fontsize{8}{9}\selectfont
\makebox[\linewidth][l]{%
  \hspace*{0.1105\linewidth}\makebox[0pt][c]{Ground Truth}%
  \hspace*{0.161\linewidth}\makebox[0pt][c]{Single}%
  \hspace*{0.161\linewidth}\makebox[0pt][c]{ERMVP}%
  \hspace*{0.161\linewidth}\makebox[0pt][c]{CoSDH}%
  \hspace*{0.161\linewidth}\makebox[0pt][c]{WhisperNet}%
  \hspace*{0.161\linewidth}\makebox[0pt][c]{Ours}}
\vspace{-3mm}
\caption{Qualitative comparison with prior methods on dense and compact scenes. Green: GT; red/orange: predictions.}
\label{fig:3d_vis}
\end{figure*}

\subsection{Ablation Studies}
\label{sec:ablations}
\begin{table}[t]
\centering
\small
\caption{Selection-policy ablation at 64~KiB on OPV2V and V2XSet (AP@0.7). Multiplicative is our full gate.}
\label{tab:policy-ap}
\begin{tabular*}{\linewidth}{@{\extracolsep{\fill}}lcc@{}}
\toprule
Selection policy & OPV2V & V2XSet \\
\midrule
Random                        & 0.6136 & 0.5087 \\
Demand-only                   & 0.6133 & 0.5488 \\
Supply-only                   & 0.8068 & 0.7435 \\
\textbf{Multiplicative (ours)} & \textbf{0.8081} & \textbf{0.7461} \\
\bottomrule
\end{tabular*}
\end{table}

\begin{table}[t]
\centering
\small
\setlength{\tabcolsep}{3pt}
\caption{Comparison of consensus designs on OPV2V. Entries are AP@0.5/AP@0.7; comm\_vol is the bandwidth exponent and Winner is decided by AP@0.7.}
\label{tab:crossover}
\resizebox{\linewidth}{!}{%
\begin{tabular}{ccccl}
\toprule
comm\_vol & $k$ & Multiplicative & Residual & Winner \\
\midrule
14 & 16   & 0.8641\,/\,0.7534 & 0.8591\,/\,0.7575 & Residual \\
15 & 32   & 0.8911\,/\,0.7886 & 0.8881\,/\,0.7919 & Residual \\
16 & 64   & \textbf{0.9034\,/\,0.8081} & 0.9009\,/\,0.8081 & Tie \\
17 & 128  & 0.9072\,/\,0.8123 & 0.9030\,/\,0.8117 & \textbf{Mult.} \\
18 & 256  & 0.9088\,/\,0.8145 & 0.9046\,/\,0.8135 & \textbf{Mult.} \\
19 & 512  & 0.9106\,/\,0.8165 & 0.9060\,/\,0.8131 & \textbf{Mult.} \\
20 & 1024 & 0.9109\,/\,0.8171 & 0.9074\,/\,0.8126 & \textbf{Mult.} \\
\bottomrule
\end{tabular}%
}
\end{table}

\noindent\textbf{Ablation Studies on Core Components.}
Table~\ref{tab:components} isolates each design choice at 64~KiB and reports AP@0.7. (1) Retraining mask-before-refine gives $0.7105$ with dense support; the inference-only swap drops to $0.5198$. (2) Replacing the 8-point deformable sampler with a rigid grid gives $0.6423$, and removing gate supervision gives $0.6594$. (3) Removing the bilateral gate gives $0.7435$, while removing collaboration falls to $0.5041$, close to random selection ($0.5087$). (4) The retrained row measures accuracy; the swap isolates dilation under identical weights.

\noindent\textbf{Ablation Studies on Selection Policy.}
Fig.~\ref{fig:supp-combined}(b) splits the transmitted tokens into cells on targets the ego-only model misses, cells on targets it already sees, and background cells, with solid bars for supply-only selection and hatched bars for multiplicative gating, where more ego-missed and fewer ego-visible or background cells are better. (1) As the budget shrinks, multiplicative gating moves the token mix away from ego-visible cells and toward ego-missed ones. (2) Fig.~\ref{fig:supp-combined}(c) asks how the same 64 tokens are spent: how many fall on targets the ego misses ($\mathrm{Prec}_{\mathrm{egM}}\!\uparrow$), how many fall on targets it already sees ($\mathrm{Prec}_{\mathrm{egV}}\!\downarrow$), and how many of the missed targets are covered ($\mathrm{Rec}_{\mathrm{egM}}\!\uparrow$). The full gate spends more of them where the ego is blind and fewer where it already sees: $\mathrm{Prec}_{\mathrm{egM}}$ rises from $0.197$ to $0.222$ on OPV2V and from $0.229$ to $0.257$ on V2XSet, $\mathrm{Prec}_{\mathrm{egV}}$ falls from $0.591$ to $0.501$ and from $0.518$ to $0.447$, and $\mathrm{Rec}_{\mathrm{egM}}$ stays at $0.890$/$0.892$. Nothing the ego already had is lost, and what it was missing is recovered. (3) Table~\ref{tab:policy-ap} confirms the accuracy effect at 64~KiB: $0.8081$/$0.7461$ on OPV2V/V2XSet versus $0.8068$/$0.7435$ for supply-only, with the margin widening as the budget shrinks (Fig.~\ref{fig:supp-combined}d).

\noindent\textbf{Ablation Studies on Consensus Design.}
Table~\ref{tab:crossover} compares our multiplicative gate with a residual alternative, $G_{\text{res}}^{(i)} = M_{\text{supply}}^{(i)} \odot (1 + M_{\text{demand}})$. The two differ in what the ego can express. In the residual, demand only scales a collaborator token between $1\times$ and $2\times$, so a cell the ego already observes stays in the candidate set. The residual also sums the two signals, so a strong supply can outweigh a weak demand without either side agreeing. Multiplication is a conjunction: a cell survives only if the ego needs it and the collaborator supplies it, so ego-visible evidence is removed before top-$k$ runs and every transmitted token adds what the ego lacks. The two designs therefore agree at the operating point but diverge as the budget grows, since the residual spends a rising share on redundancy it cannot veto.

\noindent\textbf{Ablation Studies on Anti-Dilation Refinement.}
The order of refinement and masking decides whether the budget is enforceable. Refinement is not pointwise: a $3\times3$ convolution spreads each active cell into its neighbourhood. Applied after masking, it replaces the discrete support with that receptive field, so the network rather than $k$ sets the active count, and later layers make the map dense. Refinement before masking makes the budget an invariant, since the mask is the last operation and exactly $k$ cells leave the collaborator at any depth or width. The reversed order is not a bandwidth cost but a loss of what makes the budget binding, which is why it stays dense after retraining and loses accuracy at every budget (Fig.~\ref{fig:antidilation}, Fig.~\ref{fig:supp-combined}(e)).

\noindent\textbf{Analysis on Accuracy--Latency Trade-off.}
Fig.~\ref{fig:supp-combined}(h) places all Table~\ref{tab:main-comparison} methods at 64~KiB in the GPU-stage accuracy--latency plane. SEC attains the highest AP@0.7 at the lowest latency: against ERMVP and How2comm it gains $7.8$ and $7.1$ AP@0.7 points ($0.7461$ vs.\ $0.6685$/$0.6754$) at far lower latency ($9.1$ vs.\ $76.3$/$225.8$~ms), and no other entry exceeds $0.7079$ AP@0.7 or runs faster than $9.3$~ms. The remaining baselines, including CoSDH, WhisperNet, DiscoNet, and CoAlign, are therefore both less accurate and slower.

\subsection{Qualitative Results}
\label{sec:qualitative}

\noindent\textbf{Visualization of Token Selection.}
Fig.~\ref{fig:qual-token} traces a V2XSet frame with 30 GT objects and four collaborators at $k{=}64$. Multiplication makes transmission a conjunction: a cell survives only if the ego needs it (b) and a collaborator can supply it (c), so the product (d) drops ego-visible cells before top-$k$ selection and the 64 tokens (e) land on the ego-missed targets. In this frame, detection (f) recovers 10 of the 13 objects that the ego-only model misses, and matches the unconstrained 1024~KiB output at $1/16$ the payload.

\noindent\textbf{Visualization of Sparsity and Method Comparison.}
Fig.~\ref{fig:antidilation} isolates the ordering under fixed weights. Since the mask comes last, the support stays at exactly 64 cells after every layer; refinement first lets the convolution widen it to $8{,}448$, so the budget no longer bounds the payload. Fig.~\ref{fig:3d_vis} shows the same effect at 64~KiB on two scenes. Tokens spent on cells the ego already sees or on background leave distant vehicles undetected, whereas SEC places its tokens on the right targets.

\section{Conclusion}
\label{sec:conclusion}
This paper has introduced SEC, a novel bandwidth-adaptive framework that replaces sender-driven pruning with bilateral demand-supply consensus. By systematically coupling a single-round multiplicative gate with an anti-dilation selection pipeline, SEC strictly enforces exact token budgets while precisely isolating critical ego-missed information. Comprehensive evaluations demonstrate that SEC substantially outperforms existing methods under stringent communication constraints.

{\small
\bibliographystyle{ieeenat_fullname}
\bibliography{11_references}
}


\end{document}